\documentclass[10pt,journal,compsoc]{IEEEtran}
\ifCLASSOPTIONcompsoc
  \usepackage[nocompress]{cite}
\else
  \usepackage{cite}
\fi
\ifCLASSINFOpdf
  \usepackage[pdftex]{graphicx}
\else
\fi
\usepackage{algorithm}
\usepackage{algorithmic}
\usepackage{amsmath}
\usepackage{amssymb}
\usepackage{amsthm}
\usepackage{tikz}

\usepackage{booktabs}
\usepackage{threeparttable}

\usepackage{array}
\usepackage{url,multirow,morefloats,floatflt,cancel,tfrupee}
\newtheorem{dfn}{Definition}

\begin{document}
\title{K-Core based Temporal Graph Convolutional Network for Dynamic Graphs}

\author{Jingxin Liu, %~\IEEEmembership{Member,~IEEE,}
        Chang Xu, %~\IEEEmembership{Fellow,~OSA,}
        Chang Yin, %~\IEEEmembership{Life~Fellow,~IEEE}% <-this % stops a space
        Weiqiang Wu and You Song

\IEEEcompsocitemizethanks{\IEEEcompsocthanksitem Jingxin Liu, Chang Xu and You Song are with the School of Software, Beihang University, Beijing 100191, China.\protect\\
E-mail: \{jhljx, xu\_chang, songyou\}@buaa.edu.cn

\IEEEcompsocthanksitem Chang Yin and Weiqiang Wu are with HuaRong RongTong (Beijing) Technology Co., Ltd.\protect\\
E-mail: {frank\_yinchang@163.com, Weiqiang.x.wu@foxmail.com}

\IEEEcompsocthanksitem Weiqiang Wu and You Song are corresponding authors.}% <-this % stops an unwanted space
\thanks{Manuscript accepted Oct, 2020. \\ Reference doi is http://dx.doi.org/10.1109/TKDE.2020.3033829}}


\markboth{Journal of \LaTeX\ Class Files,~Vol.~14, No.~8, August~2020}%
{J Liu, \MakeLowercase{\textit{et al.}}: K-Core based Temporal Graph Convolutional Network for Dynamic Graphs}



\IEEEtitleabstractindextext{%
\begin{abstract}
Graph representation learning is a fundamental task in various applications that strives to learn low-dimensional embeddings for nodes that can preserve graph topology information. However, many existing methods focus on static graphs while ignoring evolving graph patterns. Inspired by the success of graph convolutional networks(GCNs) in static graph embedding, we propose a novel k-core based temporal graph convolutional network, the CTGCN, to learn node representations for dynamic graphs. In contrast to previous dynamic graph embedding methods, CTGCN can preserve both local connective proximity and global structural similarity while simultaneously capturing graph dynamics. In the proposed framework, the traditional graph convolution is generalized into two phases, feature transformation and feature aggregation, which gives the CTGCN more flexibility and enables the CTGCN to learn connective and structural information under the same framework. Experimental results on 7 real-world graphs demonstrate that the CTGCN outperforms existing state-of-the-art graph embedding methods in several tasks, including link prediction and structural role classification. The source code of this work can be obtained from \url{https://github.com/jhljx/CTGCN}.
\end{abstract}

}

\maketitle

\IEEEdisplaynontitleabstractindextext

% \fi
\IEEEpeerreviewmaketitle

\IEEEraisesectionheading{\section{Introduction}\label{sec:introduction}}

\IEEEPARstart{G}{raph} embedding is a method that maps nodes into a dense and low-dimensional embedding space to preserve the topological structure, node content and other side information \cite{cui2018survey}. Recently, graph embedding has attracted considerable interest and has been proven to be an effective way to capture and preserve graph properties in a wide variety of applications, including node classification \cite{yao2019graph}, link prediction \cite{yu2017link} and anomaly detection \cite{yu2018netwalk}.

Generally, graph embedding methods can be categorized into traditional graph embedding methods and graph neural networks(GNNs). Traditional graph embedding methods often leverage random walks \cite{perozzi2014deepwalk, ribeiro2017struc2vec}, matrix factorizations \cite{yang2015network} and autoencoders \cite{wang2016structural} to generate node embeddings. Graph neural networks often aggregate neighboring features for each node guided by the graph structure \cite{zhou2018graph}. 

\begin{figure}[htbp]
\centering
\includegraphics[width=0.5\textwidth]{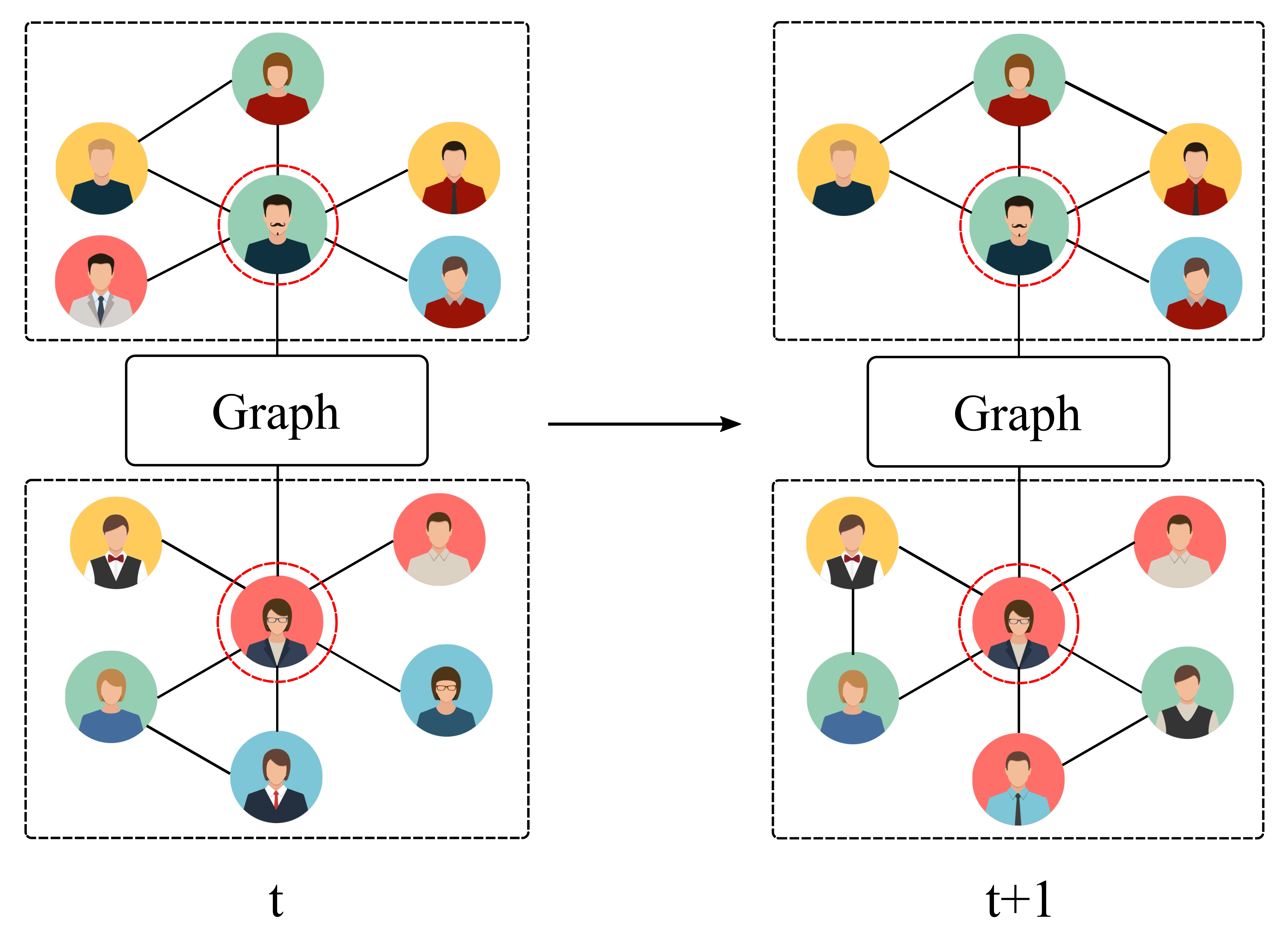}
\caption{An anomaly detection example in a dynamic communication graph under the financial fraud mining scenario. People with red circle are potential financial frauds (anomalies). }
\label{fig:1}
\end{figure}

A graph convolutional network(GCN) is a representative example of a GNN and has gained popularity in recent years due to its promising embedding performance \cite{kipf2016semi, velivckovic2017graph, hamilton2017inductive}. However, a GCN only preserves local connective proximity and suffers from oversmoothing problem when graph convolutional layers increase. Accordingly, a GCN is less capable of preserving global structural similarity in graphs. For instance, as shown in Figure \ref{fig:1}, a GCN will fail to capture the connections between those financial frauds who share similar structural roles or positions. Although DEMO-Net has attempted to retain the graph structure information, such as degree-specific information \cite{wu2019demo}, it relies on too many learnable degree-specific weight matrices, which limits its efficiency in solving large-scale graph embedding problems. 

As GCNs and related methods are less expressive at preserving global graph structures, these methods might omit some potentially essential information in graphs. As shown in Figure \ref{fig:1}, both connective proximity and structural similarity are important when mining potential financial frauds. In other words, connective proximity can help reveal financial fraud gangs in the graph, such as the people in black dashed rectangles, while structural similarity can be utilized to identify financial frauds with similar anomalous social behaviors, such as the people highlighted by red dashed circles. 

The aforementioned graph embedding methods generally focus on a given static graph. However, many real-world graphs are dynamic with the addition, deletion and changing of nodes and edges, as shown in Figure \ref{fig:1}. Various attempts have been made in recent years to overcome this challenge. However, some approaches are highly dependent on the assumption of temporal smoothness \cite{zhou2018dynamic,zhu2018high}, thereby failing to capture sharp changes during the graph evolving process. Furthermore, although some approaches \cite{yu2017link} can capture temporal evolving patterns, they have limited capacity to preserve the rich structural information in graphs. Hence, how to effectively capture the graph dynamics and preserve structural information is of vital importance and remains unresolved.

%Learning graph representations faces two main challenges. The first challenge is how to sufficiently capture the graph topology. Most existing graph embedding methods focus on either connective proximity \cite{grover2016node2vec} or structural similarity \cite{ribeiro2017struc2vec}. 

%Although SNS \cite{lyu2017enhancing} has already taken both connective proximity and structural similarity into account, it is still a static graph embedding method and is less able to capture highly non-linear graph structure.

Therefore, we want to ask the following question: \textit{is it possible to extend the properties of a GCN to make it capable of preserving local connective proximity and global structural similarity while simultaneously capturing graph dynamics?}

To answer the above question, we propose a novel k-core based temporal GCN, namely, the CTGCN, to preserve both connective proximity and structural similarity under a unified framework. The proposed method is efficient in learning node representations for dynamic graphs. As the k-core number(degeneracy) \cite{seidman1983network} is an important graph centrality and k-core decomposition \cite{matula1983smallest} naturally extracts subgraphs at multiple different scales, we design a novel k-core based graph convolutional layer to capture the topological and hierarchical properties of graphs within these nested k-core subgraphs. Most importantly, we present a generalized GCN framework that generalizes the traditional graph convolution operation into feature transformation and feature aggregation phases. Furthermore, this framework provides greater flexibility to design GCN models and enables the CTGCN to preserve both connective and structural information. Empirically, extensive experiments are conducted on 7 real-world graphs for several tasks. The results show that the CTGCN achieves gains over baselines and demonstrate that the CTGCN can learn node representations efficiently in large-scale dynamic graphs.

Overall, the main contributions of this paper are as follows:

\begin{itemize}
\item We present a novel temporal GCN that preserves both local connective proximity and global structural similarity in dynamic graphs under a unified framework and simultaneously captures evolving graph patterns.
\item We propose a novel k-core based graph convolutional layer that can uncover the topological and hierarchical properties of graphs.
\item We generalize traditional graph convolution operation into feature transformation and feature aggregation phases, which provides the GCN with more flexibility and efficiency.
%\item We evaluate the effectiveness and efficiency of the CTGCN for several real-world graphs for both connective proximity related applications(i.e., link prediction) and structural similarity related applications(i.e., structural role classification).
\end{itemize}

\section{Related Work}

% \subsection{Graph Convolutional Networks}
Graph convolutional networks have been widely studied in recent years, and they can be categorized into two types. One type applies the convolution operation directly in the spatial domain \cite{niepert2016learning,gao2018large}, which first rearranges the vertices into certain grid forms and then processes them by normal convolution operations. The other type applies convolution in the spectral domain based on Fourier transforms \cite{defferrard2016convolutional,kipf2016semi} of the graph. We refer to the graph convolutional network proposed by \cite{kipf2016semi} as the vanilla GCN. Since the vanilla GCN, many graph neural networks(GNNs) have been proposed, such as Graph Attention Network(GAT) \cite{velivckovic2017graph}, GraphSAGE \cite{hamilton2017inductive}, FastGCN \cite{chen2018fastgcn}, Graph Isomorphism Network(GIN) \cite{xu2018powerful} and Position-aware Graph Neural Networks(P-GNNs) \cite{you2019position}. Furthermore, GCN variants have been utilized to solve spatial-temporal problems. Most of these temporal GCNs combine GCN variants and recurrent architecture, such as Graph Convolutional Recurrent Network(GCRN) \cite{seo2018structured}, AddGraph \cite{zheng2019addgraph} and EvolveGCN \cite{pareja2019evolvegcn}.

%Graph Attention Network(GAT) \cite{velivckovic2017graph} leverages attention mechanisms to improve the performance of GCN. GraphSAGE \cite{hamilton2017inductive} samples a fixed-sized node neigborhood of each node, making GCN adapt to large-scale graphs. FastGCN \cite{chen2018fastgcn} also utilizes a sampling strategy called importance sampling to improve the efficiency of GCN. Graph Isomorphism Network(GIN) \cite{xu2018powerful} enhances the expressive power of GNN and is as powerful as the Weisfeiler-Lehman(WL) graph isomorphism test \cite{weisfeiler1968reduction}. Position-aware Graph Neural Networks(P-GNNs) \cite{you2019position} incorporates a node's position information with respect to all other nodes to yield position-aware node embeddings. 

%\subsection{Static Graph Embedding}
Except for GNN-based static graph embedding methods, other static methods often utilize random walks, matrix factorizations or autoencoders to generate node embeddings. Most of these static graph embedding methods focus on mining connectivity patterns in graphs. The most commonly preserved connectivity patterns are connective proximities that define neighborhood information in a graph from different levels of scope, such as first-order, second-order \cite{tang2015line} and high-order proximities \cite{ou2016asymmetric}. Previous studies preserving connective proximities include random walk-based algorithms such as DeepWalk \cite{perozzi2014deepwalk} and node2vec \cite{grover2016node2vec}, matrix factorization-based algorithms such as TADW \cite{yang2015network}, and deep learning-based algorithms such as SDNE \cite{wang2016structural}. Other static methods have taken structural similarity into account. For instance, struc2vec \cite{ribeiro2017struc2vec} is designed for learning node representations that preserve the structural identity of nodes.

%\subsection{Dynamic Graph Embedding}
Another branch of study relevant to this work is dynamic graph embedding. Dynamic graph embedding methods can be generally divided into two categories: structure preserving embedding and property preserving embedding \cite{cui2018survey}. Structure preserving embedding methods for dynamic graphs aim to preserve important structural information when a graph evolves. DynGEM \cite{goyal2018dyngem} utilizes an autoencoder to learn dynamic graph embeddings while preserving first-order and second-order proximities. HOPE \cite{zhu2018high} preserves high-order proximity and updates node embeddings through an acceleration technique. Matrix factorization-based methods, such as TIMERS \cite{zhang2018timers}, learn dynamic graph embeddings with the global structure preserved. In addition, property preserving embedding methods for dynamic graphs consider graph properties, such as transitivity, to better model the graph evolving process. For example, DynamicTriad \cite{zhou2018dynamic} introduces triad closure and models the triad closure process to capture important graph dynamics. Most of the aforementioned approaches consider dynamic graphs as a sequence of static graph snapshots, while some methods model the dynamic graph evolving process into a continuous time space, such as HTNE \cite{zuo2018embedding}. All the above methods consider the dynamic graph embedding problem for homogeneous graphs. Other methods focus on dynamic graph embedding for knowledge graphs \cite{goel2019diachronic,jin2019recurrent}.

\begin{figure*}[htbp]
\centering
\includegraphics[width=1\textwidth]{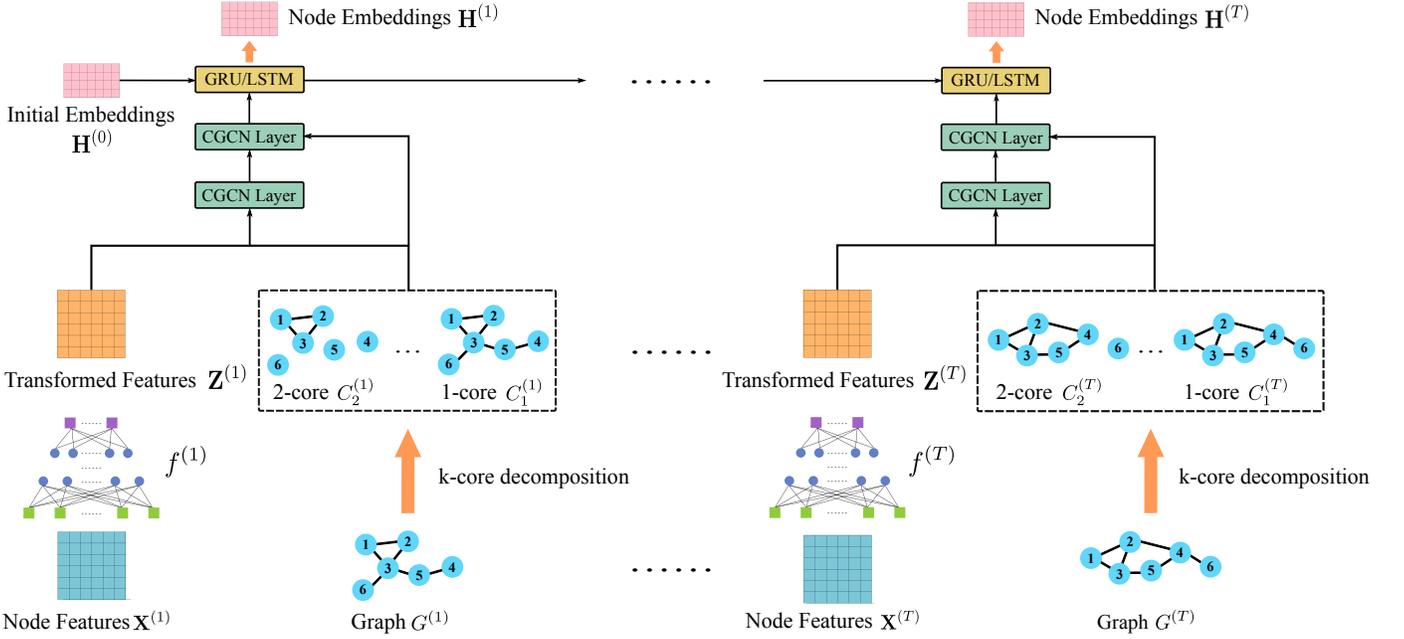}
\caption{The architecture of the k-core based temporal graph convolutional network (CTGCN). }
\label{fig:2}
\end{figure*}

\section{Preliminaries}
\subsection{Dynamic Graph Embedding} A dynamic graph at time step $t$ is an undirected graph that is defined as $G^{(t)} = (\mathcal{V}, \mathcal{E}^{(t)})$, where $\mathcal{V} = \left\{v_{1}, \dots, v_{N}\right\}$ denotes a set of $N$ nodes and $\mathcal{E}^{(t)} = \{e_{i,j}^{(t)}\}$ is a set of edges between nodes in $\mathcal{V}$. Each edge $e_{i,j}^{(t)} \in \mathcal{E}^{(t)}$ is associated with a weight $\mathbf{A}_{i,j}^{(t)} > 0$ in the adjacency matrix $\mathbf{A}^{(t)}$. Negative weighted edges are not considered in this paper.

Given a series of dynamic graph snapshots $\mathcal{G} = \left\{G^{(1)}, \cdots, G^{(T)}\right\}$, the dynamic graph embedding approach learns a sequence of mappings $F = \left\{f^{(1)}, \cdots, f^{(T)}\right\}$, where each mapping $f^{(t)} \in F$ encodes each node $v_{i}$ in $G^{(t)}$ into an embedding space $R^{d} (d \ll N)$. The learned mappings in $F$ can preserve topology information in graphs while simultaneously capturing graph dynamics.

\subsection{Graph Convolutional Networks}
Graph convolutional networks have achieved promising performance in various tasks. Formally, a single graph convolutional layer in the vanilla GCN is defined as

\begin{equation}
\mathbf{H} = \sigma\left(\tilde{\mathbf{D}}^{-\frac{1}{2}}\tilde{\mathbf{A}}\tilde{\mathbf{D}}^{-\frac{1}{2}}\mathbf{X}\mathbf{W}\right)
\label{eq:1}
\end{equation}
where $\mathbf{X}$ is the input feature matrix, $\tilde{\mathbf{A}} = \mathbf{A} + \mathbf{I}$ is the adjacency matrix with self-loop, $\tilde{\mathbf{D}}$ is a diagonal matrix with $\tilde{\mathbf{D}}_{ii} = \sum_{j=1}^{N} \tilde{\mathbf{A}}_{ij}$, $\mathbf{W}$ is a learnable weight matrix, $\sigma(\cdot)$ is a nonlinear activation function and $\mathbf{H}$ is the output embedding matrix.

Obviously, the graph convolutional layer is similar to the fully connected layer, except for the diffusion matrix $\widehat{\mathbf{A}} = \tilde{\mathbf{D}}^{-\frac{1}{2}}\tilde{\mathbf{A}}\tilde{\mathbf{D}}^{-\frac{1}{2}}$ which propagates the features of a node's neighbors to this node. In essence, graph convolution is a special form of Laplacian smoothing that computes the new features of a node as the weighted average of itself and its neighbors\cite{li2018deeper}. A one-layer GCN can preserve first-order proximity, and a multilayer GCN is often utilized in practice because it can capture high-order proximity information. However, due to the intrinsic Laplacian smoothing property, GCN suffers from oversmoothing and overfitting when graph convolutional layers increase. These problems reflect the inability of a GCN to explore global graph structure. Hence, a GCN normally includes at most 3 graph convolutional layers in practice.

\section{The Proposed Method}
\subsection{Generalized GCN Framework}
Inspired by the graph convolution operation in Equation \ref{eq:1}, we generalize the framework into two phases: feature transformation and feature aggregation, which are given by

\begin{align}
\mathbf{Z} &= f\left(\mathbf{X}\right) \\
\mathbf{H} &= h\left(g\left(\mathbf{A}\right), \mathbf{Z}\right)
\label{eq:2}
\end{align}
where $\mathbf{X}$ is the input feature matrix, $f$ is the feature transformation function that maps $\mathbf{X}$ into a latent space, $\mathbf{A}$ is the adjacency matrix,  $g$ is a function of $\mathbf{A}$ that defines the propagation rule, $h$ is the feature aggregation function that aggregates features for each node based on the propagation rule $g$, and $\mathbf{H}$ is the embedding matrix that integrates the information from feature transformation and graph structure.

Many previous works can be included in this framework, as summarized in Table \ref{table:1}. We note that most GCNs define a linear transformation as $f$, while the PPNP method utilizes a multilayer perceptron(MLP) to model a nonlinear $f$. Although these methods have different $f$ and $g$ combinations, they all choose the outputs of $h$ as the final embeddings in practice. We argue that both the outputs of $f$ and $h$ can be utilized as node representations, which will be discussed later. By generalizing the traditional graph convolution operation into feature transformation and feature aggregation, we provide more flexibility to design different kinds of GCNs.

\begin{table}[htbp]
\caption{Summary of previous GCNs under the generalized GCN framework.}
\label{table:1}
\centering
\begin{threeparttable}
\begin{tabular}{cccc}
\hline
\specialrule{0em}{1pt}{1pt}
Method & $f$ & $g$ & $h$ \\
\specialrule{0em}{1pt}{1pt}
\hline
\specialrule{0em}{1pt}{1pt}
GCN \cite{kipf2016semi} & $\mathbf{X}\mathbf{W}$  & $\widehat{\mathbf{A}}$ & $\sigma\left(g(\mathbf{A})f(\mathbf{X})\right)$ \\
\specialrule{0em}{1pt}{1pt}
SGC \cite{wu2019simplifying} & $\mathbf{X}\mathbf{W}$ & $\widehat{\mathbf{A}}^{k}$ & ${\rm softmax}\left(g(\mathbf{A})f(\mathbf{X})\right)$ \\
\specialrule{0em}{1pt}{1pt}
MixHop \cite{abu2019mixhop}    & $\mathbf{X}\mathbf{W}$  & $\widehat{\mathbf{A}}^{k} $ & $\bigg\Vert_{k \in \mathcal{P}}{ \; \sigma\left(g_{k}(\mathbf{A})f(\mathbf{X})\right)}$ \\
\specialrule{0em}{1pt}{1pt}
GraphHeat \cite{xu2019graph} & $\mathbf{X}\mathbf{W}$ & $e ^{-\lambda \mathbf{L}} + \mathbf{I} $ \tnote{*} & $\sigma\left(g(\mathbf{A})f(\mathbf{X})\right)$ \\
\specialrule{0em}{1pt}{1pt}
PPNP \cite{klicpera2018predict} & $MLP(\mathbf{X})$  & $g(\mathbf{A}) \tnote{**}$ & $\sigma(g(\mathbf{A})f(\mathbf{X}))$ \\
\specialrule{0em}{1pt}{1pt}
\hline
\specialrule{0em}{1pt}{1pt}
\end{tabular}
\begin{tablenotes}
\footnotesize
\item[*] $\mathbf{L} = \mathbf{I} - \mathbf{D}^{-\frac{1}{2}}\mathbf{A}\mathbf{D}^{-\frac{1}{2}}$
\item[**] $g(\mathbf{A}) = \alpha(\mathbf{I} - \left(1 - \alpha\right)\widehat{\mathbf{A}})^{-1}$
\end{tablenotes}
\end{threeparttable}
\end{table}

\subsection{Core-based Temporal GCN}
We propose a novel k-core based temporal graph convolutional network, namely, the CTGCN, to preserve both connective proximity and structural similarity in dynamic graphs. As shown in Figure \ref{fig:2}, the CTGCN includes two modules: a core-based GCN module and a graph evolving module. Note that this is also an instance of a generalized GCN framework. Although the CTGCN is a temporal GCN method, we can regard $h$ as a complex spatiotemporal feature aggregation function.

\subsubsection{Core-based GCN Module}
To capture rich graph structure information, we leverage k-core subgraphs to design a core-based GCN, namely, a CGCN. Here, we first give the definition of k-core \cite{nikolentzos2018degeneracy}.

\begin{dfn}[k-core]
Consider a graph $G$ and a subgraph $G'$ of the graph $G$. $G'$ is a k-core of $G$ if $G'$ is a maximal subgraph of $G$ in which all nodes have a degree of at least k.
\end{dfn}

From the definition above, we can infer some interesting k-core properties. One property is that k-core numbers of nodes have a strong positive correlation with node degrees in real-world graphs \cite{shin2016corescope}. Another property is that k-cores form a nested structure. Formally, let $\mathcal{C} = \left\{C_{0}, C_{1}, \cdots, C_{k_{max}}\right\}$ be the k-core set of a graph $G$; then, all k-cores in $C$ form a nested chain, which is given by

\begin{equation}
C_{k_{max}} \subseteq \cdots C_{1} \subseteq C_{0} = G
\label{eq:4}
\end{equation}
where $k_{max}$ is the max k-core number of $G$. All k-cores can be extracted by k-core decomposition algorithms, which have a linear time complexity with the edge number \cite{batagelj2003m} and are efficient at processing large-scale graphs. It is worth noting that k-core decomposition can be regarded as a kind of graph partitioning strategy, and a similar graph partitioning strategy has been leveraged in ClusterGCN \cite{chiang2019cluster}. ClusterGCN partitions a graph into several subgraphs by graph clustering algorithms to restrict the neighbor expansion in the GCN layer and hence prevent the oversmoothing problem in a GCN. However, our aim is to augment the GCN's properties, allowing the CTGCN not only to preserve local connective proximity, but also to capture global structural information in dynamic graphs.

\begin{figure}[htbp]
\centering
\includegraphics[width=0.48\textwidth]{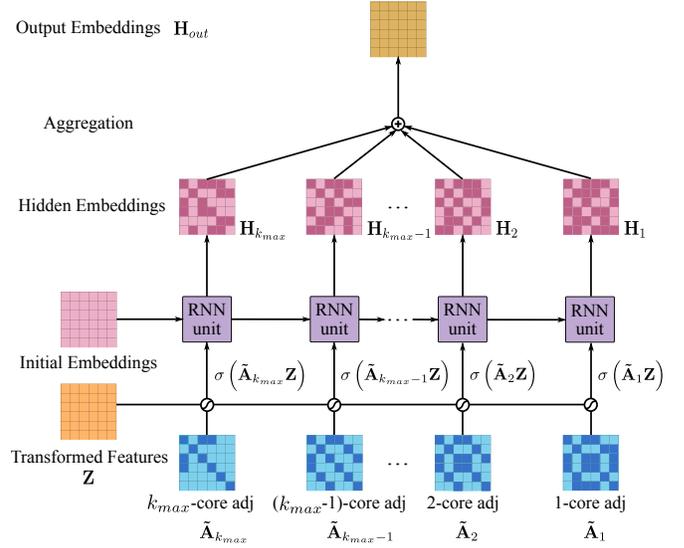}
\caption{The architecture of the CGCN layer. }
\label{fig:3}
\end{figure}

As the k-core naturally defines a hierarchical nested subgraph structure, all k-cores can be utilized to propagate node features at different scales in a graph. The definition of the core-based graph convolutional layer is illustrated in Figure \ref{fig:3}. Let $\mathbf{Z} = f\left(\mathbf{X}\right)$ be the feature transformation matrix. The feature propagation on each k-core subgraph $C_{k}$ is defined as:

\begin{equation}
\mathbf{Z'}_{k} = \sigma\left(\tilde{\mathbf{A}}_{k}\mathbf{Z}\right)
\label{eq:5}
\end{equation}
where $\tilde{\mathbf{A}}_{k}$ is the expanded adjacency matrix of the k-core subgraph with added self-connections for all nodes, in order to make all k-core adjacency matrices have the same dimension. 

After feature propagation on k-core subgraphs, an intuitive approach to obtain more expressive feature matrices is to aggregate all $\mathbf{Z'}_{k}$ by aggregation functions, such as {\sc Sum}. We demonstrate the effectiveness of this intuitive approach in later sections. Furthermore, we assume that each graph has a latent k-core subgraph evolution process. Recurrent neural networks(RNNs), such as Gated Recurrent Unit(GRU) \cite{cho2014learning} or Long Short-Term Memory(LSTM) \cite{hochreiter1997long}, can be employed to capture complex subgraph evolution patterns in the latent k-core subgraph evolution process, which is defined as:

\begin{equation}
\mathbf{H}^{l} = {RNN}(\mathbf{Z'}_{k_{max} + 1 - l}, \mathbf{H}^{l - 1}), \quad l=1,\cdots, k_{max}
\label{eq:6}
\end{equation}
where $\mathbf{H}^{0} = \mathbf{0}$ is a zero matrix. The output of the CGCN layer is given by:

\begin{equation}
\mathbf{H}_{out} = \sum_{l = 1} ^{k_{max}} \mathbf{H}^{l}
\label{eq:7}
\end{equation}

As shown in Figure \ref{fig:3}, the feature transformation matrix $\mathbf{Z}$ is shared across all RNN units, which reduces the number of learnable weight matrices. We also note that $\tilde{\mathbf{A}}_{k + 1}$ is involved in $\tilde{\mathbf{A}}_{k}$. Hence, changing matrices $\Delta\tilde{\mathbf{A}}_{k} = \tilde{\mathbf{A}}_{k} - \tilde{\mathbf{A}}_{k + 1}$ are utilized for sparse matrix multiplication, which reduces memory cost and improves computational efficiency. Similar to the vanilla GCN layer, the CGCN layer only preserves first-order proximity. Multiple CGCN layers are helpful to capture connective information in a larger receptive field.

\subsubsection{Graph Evolving Module}
To effectively capture graph dynamics during the graph evolving process, we utilize an RNN, such as a GRU or LSTM, to model the temporal dependency among node representations of CGCNs at different timestamps. Note that this module can also be augmented by other RNN architectures if more complex temporal features are needed. The temporal dependency of k-core subgraphs across different timestamps can also be considered, but this will affect the computational efficiency of the CTGCN. 

Formally, the graph evolving module is defined as follows:

\begin{equation}
\mathbf{H}^{(t)} = {RNN}(CGCN^{(t)}(\mathbf{A}^{(t)}, \mathbf{X}^{(t)}), \mathbf{H}^{(t - 1)})
\label{eq:8}
\end{equation}
where $\mathbf{H}^{(0)} = \mathbf{0}$ is a zero matrix, $\mathbf{A}^{(t)}$ is the adjacency matrix of $G^{t}$, $\mathbf{X}^{(t)}$ is the node feature matrix of $G^{t}$ and $t = 1, \cdots, T$. Each CGCN generates static node embeddings for the RNN unit, and the RNN unit decides how much information should be preserved in the layer of the next timestamp.

\subsection{Connective Proximity Preserving CTGCN}
To preserve connective proximity in dynamic graphs, we follow previous GCNs to utilize the outputs of the $h$ function as final embeddings. In connective proximity preserving CTGCN(CTGCN-C), $f$ is denoted as a linear mapping, that is, $\mathbf{Z} = \mathbf{X}\mathbf{W} + \mathbf{b}$. For each $CGCN^{(t)}$, we employ two CGCN layers to capture the connective information within 2-hop neighbors.

To this end, the overall objective of the CTGCN-C is summarized below:

\begin{align}
\label{eq:9}
\mathcal{L}_{c} &= \sum_{t=1}^{T} \sum_{u \in \mathcal{V}} \mathcal{L}_{u}^{t} \\
\label{eq:10}
\mathcal{L}_{u}^{t} &= \sum_{v \in \mathcal{N}_{\rm w}^{t}(u)} - {\rm log}\,\left(F_{p}\right) -Q \cdot \sum_{v' \in P_{n}^{t}(u)} {\rm log}\left(1-F_{n} \right)
\end{align}
where $F_{p} = \sigma\left(\left<\mathbf{h}_{u}, \mathbf{h}_{v}\right>\right)$, $F_{n} = \sigma\left(\left<\mathbf{h}_{u}, \mathbf{h}_{v'}\right>\right)$, $\mathcal{N}_{\rm w}^{t}(u)$ is the set of nodes that co-occur with $u$ on fixed-length random walks, $P_{n}^{t}$ is a negative sampling distribution which is usually a function of node degrees, and $Q$ is a hyperparameter used to balance the positive and negative samples.

Through this unsupervised loss function, nodes co-occurring in fixed-length random walks are encouraged to have similar representations. Hence, the learned node representations can preserve local connective proximity in dynamic graphs.

\subsection{Structural Similarity Preserving CTGCN}
In previous works, structural similarity preserving graph embedding methods can be summarized into two categories: one uncovers structural similarity based on the structural similarity graph, which is derived from the original graph(i.e., struc2vec), and the other encodes the neighboring structure into embeddings(i.e., DEMO-Net). We focus on the second method in this work. Formally, we first provide more accurate definitions to describe structural similarity \cite{tu2018deep}.

\begin{dfn}[structural equivalence] Consider two nodes $u$ and $v$ in a graph; $u$ and $v$ are structurally equivalent if and only if $\mathcal{N}_{u} = \mathcal{N}_{v}$, where $\mathcal{N}_{u}$ and $\mathcal{N}_{v}$ are the neighbor sets of u and v, respectively.
\end{dfn}

Many structural similarity measures are developed based on structural equivalence. A common similarity measure is the Jaccard index \cite{jaccard1901etude}, which quantifies the ratio of common neighbors between two nodes. We can note that these measures are only local measures, which restricts similar nodes to sharing the same neighbor set. 

To capture the equivalence relationship between nodes that occupy the same structural role in a graph without having common neighbors, regular equivalence relaxes the condition of structural equivalence.

\begin{dfn}[regular equivalence] Consider two nodes $u$ and $v$ in a graph; $u$ and $v$ are regularly equivalent if and only if $\mathcal{N}_{u}$ and $\mathcal{N}_{v}$ are regular equivalent.
\end{dfn}

The definition of regular equivalence is recursive, because one needs to identify the equivalence relationship of the neighbors of two nodes before one can identify the equivalence relationship of the two nodes. Common regular equivalence-based structural similarity measures include SimRank \cite{jeh2002simrank} and Vertex Similarity(VS) \cite{leicht2006vertex}. The node similarity matrix defined by the VS measure is an infinite power series of the adjacency matrix, and highly linked node pairs tend to have greater structural similarity values. According to the k-core definition, we note that nodes in high-order cores are densely connected, which means that node pairs in high-order cores are more structurally similar than node pairs in low-order cores. As CGCN layers restrict feature aggregation in each core, nodes in the same core tend to have similar node representation, which enables the CTGCN to preserve global structural similarity in dynamic graphs.

% In previous works, SDNE \cite{wang2016structural} utilizes a deep autoencoder to encode each row of an adjacency matrix into a latent embedding space, which preserves structural equivalence in graphs. Furthermore, DRNE \cite{tu2018deep} encodes the neighboring information to make nodes with similar neighbors have similar embeddings, which preserves regular equivalence in graphs.

In structural similarity preserving CTGCN(CTGCN-S), $f$ is denoted as a nonlinear mapping, that is $\mathbf{Z} = MLP(\mathbf{X})$. For each $CGCN^{(t)}$, we only utilize 1 CGCN layer to aggregate neighboring features. The overall objective of the CTGCN-S is summarized below:

\begin{equation}
\label{eq:11}
\mathcal{L}_{s} = \sum_{t=1}^{T} \lVert \mathbf{Z}^{(t)} - \mathbf{H}^{(t)}\rVert_{F}^{2} \\
\end{equation}
where $\mathbf{Z}^{(t)}$ is the final node embedding matrix, and $\mathbf{H}^{(t)}$ is the $t$-th output embedding matrix of the graph evolving module.

In the initial gradient descent step, each node $u$ will aggregate the features of its first-order neighbors into $\mathbf{Z}^{(t)}_{u}$, which preserves the local structure in dynamic graphs. By optimizing the loss function in Equation \ref{eq:11} iteratively, $\mathbf{Z}^{(t)}_{u}$ will aggregate the features of high-order neighbors recursively. After multiple iterations, $\mathbf{Z}^{(t)}$ will be close to the fixed point and is able to capture the global structure information in dynamic graphs. Note that the CTGCN-S can also combine the loss functions in Equation \ref{eq:9} and Equation \ref{eq:11} to preserve both local proximity and global structure information.

On the other hand, the optimization of Equation \ref{eq:11} can also be regarded as a knowledge distillation process \cite{hinton2015distilling}. Thanks to the universal approximation theorem \cite{hornik1991approximation}, graph information in the complex feature aggregation neural network ${\rm Net}_{h}$ is distilled into the simple feature transformation neural network ${\rm Net}_{f}$, which can accelerate the model evaluation efficiency because only ${\rm Net}_{f}$ is needed for generating test node embeddings. This implicit knowledge distillation strategy might also be leveraged for interactive systems in which the latent interaction graph is almost fixed, but node features change dynamically \cite{kipf2018neural}. Hence, we can note that our proposed generalized GCN framework not only includes many previous methods, as shown in Table \ref{table:1}, but also has potential connections with the knowledge distillation when Equation \ref{eq:11} is used, which demonstrates that the generalized GCN framework provides the GCN with more flexibility and efficiency.

\subsection{Discussion}
\label{discussion}

We compare the proposed CTGCN with existing GCN-based methods regarding the graph properties, as illustrated in Table \ref{table:2}.

\begin{table}[htbp]
\caption{Comparison with existing GCN-based methods.}
\label{table:2}
\centering
\begin{tabular}{cccc}
\hline
Method & local proximity & global structure & dynamics \\
\hline
\specialrule{0em}{1pt}{1pt}
GCN \cite{kipf2016semi} & $\checkmark$  & $\times$ & $\times$ \\
GAT \cite{velivckovic2017graph} & $\checkmark$  & $\times$ & $\times$\\
GCRN \cite{seo2018structured} & $\checkmark$ & $\times$ & $\checkmark$ \\
EvolveGCN \cite{pareja2019evolvegcn}  & $\checkmark$ & $\times$ & $\checkmark$ \\
CTGCN &  $\checkmark$  & $\checkmark$ & $\checkmark$  \\
\hline
\end{tabular}
\end{table}

We note that the existing GCN-based methods shown in Table \ref{table:2} do not preserve all three properties. Most importantly, these methods can only preserve connective proximity in graphs. To improve the expressive power of GCN, we extend the vanilla GCN layer into the core-based GCN layer, making the CTGCN preserve both local connective proximity and global structural similarity in dynamic graphs. Although the CTGCN leverages the RNN to capture dynamic graph evolving patterns such as GCRN and EvolveGCN, the CTGCN also considers the latent k-core subgraph evolution process in dynamic graphs. Furthermore, the CTGCN has a wider range of applications than GCRN and EvolveGCN.

\subsection{Complexity Analysis}
\label{comp_ana}
To obtain all k-cores in dynamic graphs, we utilize a k-core decomposition algorithm that is linear to the edge number\cite{batagelj2003m}. Let $\left|\mathcal{E}\right|_{max}$ be the maximal edge number across all dynamic graphs, then the time complexity of extracting k-cores is $O\left(T\left|\mathcal{E}\right|_{max}\right)$ and can be reduced to $O\left(\left|\mathcal{E}\right|_{max}\right)$ if parallel computing techniques are utilized.

For the CGCN module, we consider that both adjacency matrices and input node features are stored as sparse matrices. Then, the time complexity of the CGCN module is $O(Tlk_{max}\left|\mathcal{E}\right|_{max})$, where $l$ is the layer number of the CGCN layers. Here, the time complexities of the RNNs in the CGCN are not considered, as their time complexities are lower than those of sparse matrix multiplications. Therefore, the time complexity of feature aggregation in the CGCNs is linear to the edge number.

For the graph evolving module, the time complexity is equal to that of a GRU(or LSTM). As $CGCN^{(t)}(\mathbf{A}^{(t)}, \mathbf{X}^{(t)})$ and $\mathbf{H}^{(t - 1)}$ are inputs of the RNN model, assume they are $N \times d$ and $N \times d'$ matrices, respectively. Then, the time complexity is $O(k'N(d + d'))$, where $k'$ is a constant independent of $N$. Overall, the time complexity of the CTGCN is linear to the node number, which indicates that our proposed method can be applied to large-scale graph settings.

\section{Experiments}
In this section, we evaluate the proposed method in the dynamic setting from four aspects: the performance on the connective proximity-related tasks, the structural similarity-related tasks, the parameter sensitivity and the time cost. Experimental analyses are presented as follows.

\subsection{Experimental Setup}
\label{exp_setup}

\noindent\textbf{Baselines} We employ the following graph embedding methods as baselines:
\begin{itemize}
% \item DeepWalk \cite{perozzi2014deepwalk}. It is a static graph embedding model combining truncated random walk with SkipGram model \cite{mikolov2013efficient} to learn node representations.
% \item node2vec \cite{grover2016node2vec}. It designs a biased random walk procedure to explore diverse neighbourhoods and utilizes SkipGram model to learn static node embeddings.
\item struct2vec \cite{ribeiro2017struc2vec}. It constructs a hierarchical graph to measure structural similarity at different scales and leverages the SkipGram model \cite{mikolov2013efficient} to learn structural representation for nodes in a static graph.
\item GCN \cite{kipf2016semi}. It encodes the graph structure directly using a neural network model based on a simple and well-behaved layerwise propagation rule. %, which is a localized first-order approximation of spectral graph convolutions.
\item GAT \cite{velivckovic2017graph}. It is a variant of the GCN, which leverages masked self-attentional layers to aggregate neighborhoods' features with different weights.

\item GIN \cite{xu2018powerful}. It replaces linear mappings in the GCN with MLPs and adds parameters to enhance embeddings for each node itself when aggregating neighborhood information.

\item CGCN-C. It is the static version of the CTGCN-C without the graph evolving module. The output of the CGCN layers is utilized as the final node embedding.

\item CGCN-S. It is the static version of the CTGCN-S without the graph evolving module. The output of nonlinear feature transformation layers is utilized as the final node embedding.

\item DynGEM \cite{goyal2018dyngem}. It extends the SDNE into dynamic graphs and leverages deep autoencoders to generate dynamic graph embedding incrementally.

\item dyngraph2vec \cite{goyal2020dyngraph2vec}. It utilizes historical adjacent matrices to reconstruct the current adjacent matrices through the encoder-decoder architecture. There are three variants: dynAE, dynRNN and dynAERNN.

%\item TIMERS \cite{zhang2018timers}. It utilizes singular value decomposition(SVD) to incrementally generate dynamic graph embedding and automatically restart the SVD when accumulated error exceeds the preset error threshold.

\item GCRN \cite{seo2018structured}. It combines the GCN model with an RNN to generate node embeddings in dynamic graphs. We choose the GCN as the backbone of the GCRN, which is conceptually equivalent to the ChebNet \cite{defferrard2016convolutional} in the original GCRN model.

% \item VGRNN \cite{hajiramezanali2019variational}.

\item EvolveGCN \cite{pareja2019evolvegcn}. It is a dynamic variant of the GCN, which leverages an RNN to evolve GCN parameters. There are two versions of the EvolveGCN method: EvolveGCN-O and EvolveGCN-H. 
\end{itemize}

%, we perform a grid search on the values of hyper-parameters, and we choose a specific combination of them for each task on each data set, which results in the best performance regarding to the area under curve(AUC) score metric.

\noindent\textbf{Parameter Settings}
% We set window size as 10, walk length as 40 and walks per node as 20 for DeepWalk. We set p as 1 and q as 2 in node2vec to discover more neighborhood information, other parameter settings are the same as DeepWalk. 
For static GCN-based methods, we utilize 2 layers in the GCN, GAT and GIN. We also utilize 2 layers in the GCRN and EvolveGCN. For DynGEM, we set $\alpha$ as $10^{-5}$, $\beta$ as $10$, $\gamma_{1}$ as $10^{-4}$ and $\gamma_{2}$ as $10^{-4}$. We set $\beta$ as $5$, the look back as 3, $\gamma_{1}$ as $10^{-6}$ and $\gamma_{2}$ as $10^{-6}$ for dynAE, dynRNN and dynAERNN.

For the proposed method, we utilize a linear layer and 2 CGCN layers in the CGCN-C and CTGCN-C, while we utilize a 3-layer MLP and 1 CGCN layer in the CGCN-S and CTGCN-S. We set the k-core subgraph number of CGCN layers as the maximum k-core number across all dynamic graphs. We adopt the Adam optimizer \cite{kingma2014adam} with learning rate of $0.001$ and weight decay of $5 \times 10^{-4}$. We use one-hot node feature matrices for the CGCN-C and CTGCN-C, and use degree-based node feature matrices for the CGCN-S and CTGCN-S. For fair comparison, the dimensionality of embeddings $d$ is set to 128 for all compared methods.

\subsection{Link Prediction}
Given a series of dynamic graphs, the link prediction predicts the existence of an edge in the next time step $t + 1$ based on the information of embeddings in the current time step $t$. To obtain a labeled edge set from graph $G^{(t)}$, we generate random sampled edges in time step $t$ as positive examples, and generate negative samples by sampling node pairs that are not connected by any edges. Then, we follow the methodology in \cite{grover2016node2vec}  and compute edge feature vectors by utilizing the Hadamard operation between embedding vectors of node pairs in the labeled edge sets. We train a logistic regression(LR) classifier to discriminate positive and negative edge samples. The area under the curve(AUC) is utilized as the evaluation metric, and the averaged AUC scores are reported as link prediction results. Note that all static graph embedding methods are retrained on each graph snapshot to generate node embeddings, and all dynamic graph embedding methods incorporate historical graph information to yield node embeddings of the current timestamp. 

\begin{table}[htbp]
\caption{Statistics of datasets utilized for link prediction. \protect\\ ($D_{\rm max}$: maximal degree, $k_{\rm max}$: maximal k-core number.)}
\label{table:3}
\centering
%\resizebox{\textwidth}{5mm}{
\begin{tabular}{cccccc}
\hline
Dataset & \#Nodes & \#Edges & $D_{\rm max}$ & $k_{\rm max}$ & \#Snapshots \\
\hline
% SBM      & 1000   & 4870863 & 161 & 86 & 50 \\
UCI      & 1899   & 59835   & 198  & 16  & 7  \\
AS       & 6828   & 1947704 & 1458  & 13  & 100 \\
Math     & 24740  & 323357 & 231 & 15 &  77 \\
Facebook & 60730  & 607487 & 203  & 9 & 27 \\
Enron    & 87036  & 530284 & 1150  & 22 & 38 \\

\hline
\end{tabular}%}
\end{table}

We employ five datasets in this experiment, including a student message network at the University of California, Irvine\footnote{http://konect.cc/networks/opsahl-ucsocial/}, a communication network from Autonomous Systems\footnote{http://snap.stanford.edu/data/as-733.html}, a social network from the Facebook Corporation\footnote{http://networkrepository.com/fb-wosn-friends.php}, an email contact network from the Enron Corporation\footnote{http://networkrepository.com/ia-enron-email-dynamic.php} and a user interaction network from the stack exchange website Math Overflow\footnote{http://snap.stanford.edu/data/sx-mathoverflow.html}. In practice, we split the datasets by month and remove incomplete data. Then, the complete data from continue months are retained for the link prediction task. Detailed statistics of the aforementioned datasets are shown in Table \ref{table:3}.

\begin{table}[htbp]
\caption{Average AUC scores of all timestamps for link prediction.}
\label{table:4}
\centering
\begin{tabular}{ccccccc}
\hline
\specialrule{0em}{1pt}{1pt}
Method & UCI & AS & Math & Facebook & Enron \\
\hline
\specialrule{0em}{1pt}{1pt}
% DeepWalk & 0.6769  & 0.8038 & 0.7742 &   &    & 0.8367  \\
% node2vec & 0.6819 & 0.8090 & 0.7802 &    &    &   \\
GCN & 0.7729   & 0.7835 & 0.7986 & 0.6942 & 0.7879   \\ 
GAT & 0.7668   & 0.7906  & 0.8351 & 0.6991 & 0.8344  \\ 
GIN & 0.8366 & 0.8571  &  0.8681  & 0.7415 &  0.8453  \\
CGCN-C & 0.9287 & 0.9330 & 0.9145 & 0.8025 & 0.9211   \\
CGCN-S &  0.9375 & 0.9317 & 0.9119 & 0.8257 & 0.9076  \\
\hline
\specialrule{0em}{1pt}{1pt}
DynGEM &  0.9032  & 0.9372  & 0.9025 & 0.8019 & 0.8926  \\  
dynAE  & 0.9189 & 0.9404  &  0.9178  &   ---   &  ---   \\
dynRNN & 0.8927 & 0.9107  &  ---  &   ---  &   ---     \\
dynAERNN & 0.9008 & 0.9139  &  --- &  ---  &  ---   \\
% TIMERS &    &    &    & 0.6129 & 0.7707 & 0.7177 \\
GCRN   &  0.8579   & 0.8648 &   0.8217  & 0.7262  &  0.8807  \\
% VGRNN  &    xx    &   xx     &    xx    &   xx &   xx  &  xx \\
EvolveGCN & 0.9102 & 0.9227  & 0.9034 & 0.8056 & 0.9025   \\
CTGCN-C &  \textbf{0.9434}  & 0.9578 & \textbf{0.9691} & \textbf{0.8836} & \textbf{0.9769}  \\
CTGCN-S & 0.9403 & \textbf{0.9630} & 0.9266 & 0.8324 & 0.9321 \\
\hline
\end{tabular}
\end{table}

\begin{figure}[htbp]
\centering
\includegraphics[width=0.49\textwidth]{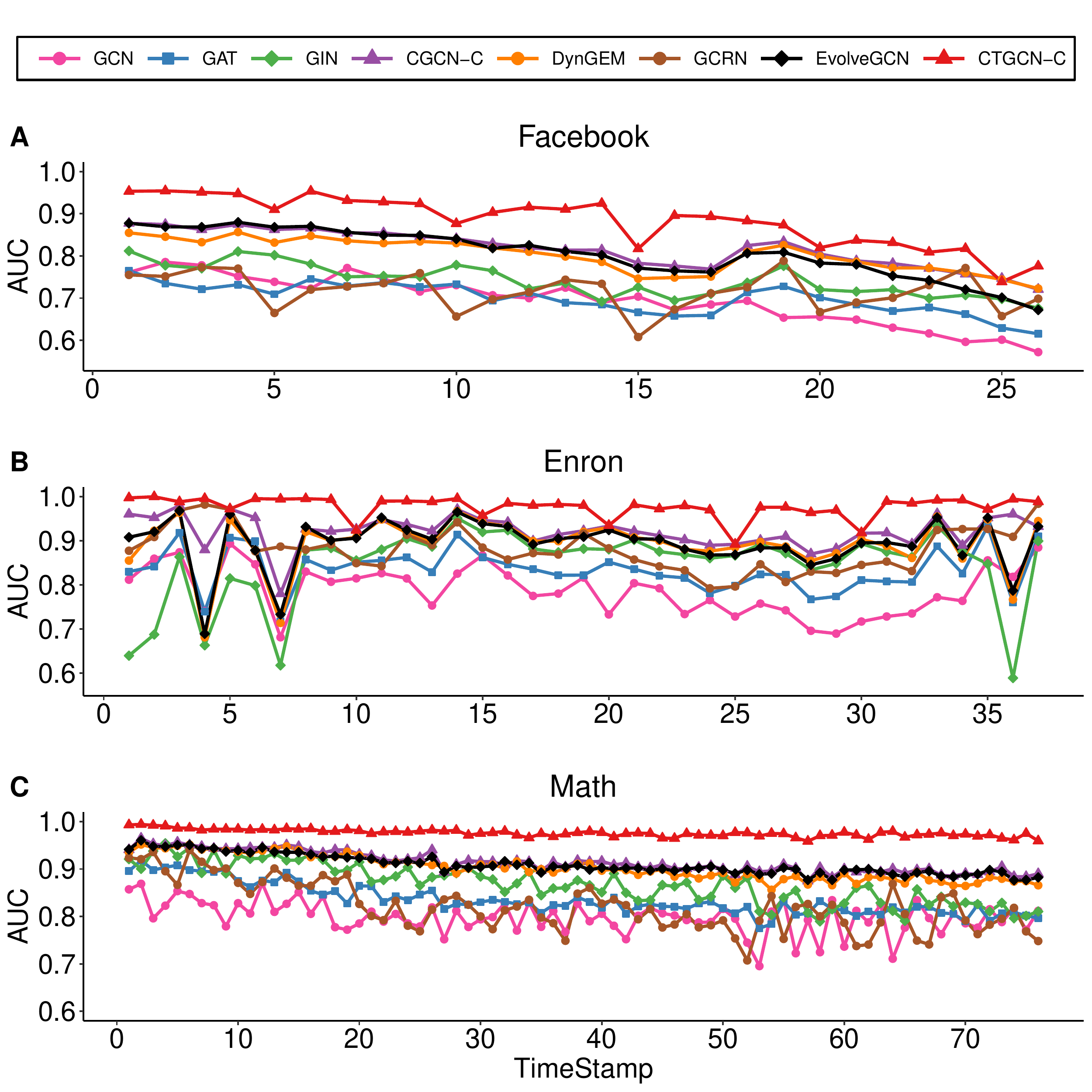}
\caption{AUC of link prediction on dynamic graphs. }
\label{fig:4}
\end{figure}

We report link prediction results on these real-world graphs where the best results are indicated in bold, as illustrated in Table \ref{table:4}. It can be observed that the proposed CTGCN-C method significantly outperforms other compared methods across all dynamic graphs. This indicates that the CTGCN-C can capture hierarchical graph structure information in dynamic graphs, which is helpful for its node representations to predict link formation and deletion. Furthermore, we observe that the CTGCN-S is also comparable to the CTGCN-C in the link prediction task. This demonstrates that the CTGCN-S is also capable of preserving local connective proximity in dynamic graphs, and feature transformation function $f$ can yield expressive node representations under the guidance of feature aggregation function $h$, as shown in Equation \ref{eq:11}. In particular, the CGCN-C and CGCN-S consistently outperform other static graph embedding methods, which suggests that the CGCN module is more helpful than the graph evolving module for the improved performance of the CTGCN in the link prediction task.

To further evaluate the stability of graph embedding methods over timestamps, we record the AUC scores of all compared methods for each timestamp, as shown in Figure \ref{fig:4}. We see that the AUC values of most approaches fluctuate sharply during the graph evolving process, while the CTGCN-C behaves more stably than other methods for most timestamps. Hence, the results further demonstrate that the CTGCN can learn stable and expressive node representations for different timestamps.

\subsection{Graph Centrality Prediction}
\label{sec:rep}

To evaluate the structural similarity preserving performance of all compared methods, we set up the graph centrality prediction on dynamic graphs. We employ the UCI and AS datasets in this task. As graph centralities have been utilized to characterize the structural role and importance of nodes, we choose four popular centralities: closeness centrality \cite{okamoto2008ranking}, betweenness centrality \cite{barthelemy2004betweenness}, eigenvector centrality \cite{bonacich2015eigenvector} and k-core centrality \cite{kitsak2010identification} as ground truths in this task. Our aim is to use node representations of all compared methods to train a ridge regression model and predict the above four centrality scores. The mean square error(MSE) is utilized as the evaluation metric.

\begin{table}[htbp]
\caption{The average MSE value of predicting centralities on the UCI dataset($*10^{-2}$).}
\label{table:5}
\centering
\begin{tabular}{ccccc}
\hline
Method & closeness & betweeness & eigenvector & k-core \\
\hline
struc2vec & 0.1995  & 0.2579 & 3.4294 & 12.3348 \\
GCN & 4.1749  & 0.4732 & 7.2990 & 232.6379 \\
GAT & 4.0186 & 0.5222 & 7.5461 & 250.0386 \\
GIN & 1.8912 & 0.1223 &  1.7228  & 87.4469 \\
CGCN-C & 0.5103 & 0.3237 & 2.6174 & \textbf{3.5643} \\
CGCN-S & \textbf{0.1830} & \textbf{0.0413} & 2.4864 & 4.5918\\
\hline
\specialrule{0em}{1pt}{1pt}
DynGEM & 1.0822  & 0.2180 & 3.3836 & 31.5852 \\
dynAE & 2.2514  & 0.2555  & 4.8254 & 79.8863 \\
dynRNN & 2.5754 & 0.2346  & 6.0414 & 72.6757 \\
dynAERNN & 1.5846  & 0.3114  & 4.9229 & 62.6733 \\
GCRN & 3.0201  & 0.4527 & 6.4607 & 154.4173 \\
EvolveGCN & 0.6344  & 0.5059 & 6.5992 & 139.9802 \\
CTGCN-C & 0.8333 & 0.3652 & 3.5197 & 7.9841 \\
CTGCN-S & 0.2351 & 0.0533 & \textbf{1.5651} & 7.0822 \\
\hline
\end{tabular}
\end{table}

\begin{table}[htbp]
\caption{The average MSE value of predicting centralities on the AS dataset($*10^{-2}$).}
\label{table:6}
\centering
\begin{tabular}{ccccc}
\hline
Method & closeness & betweeness & eigenvector & k-core \\
\hline
struc2vec & \textbf{0.1631}  & 3.5613 & 1.5077 & 29.9080 \\
GCN & 3.4562 & 4.0092 & 2.5071 & 128.1415 \\
GAT & 3.8602 & 4.7150 & 3.1137 & 174.1997 \\
GIN & 1.7279 & 0.3949 & 1.5399 &  58.8688  \\
CGCN-C & 0.2612 & 4.1666 & 1.3507 & \textbf{5.0986} \\
CGCN-S & 0.3694 & 1.1880 & 1.0412 & 22.7628 \\
\hline
\specialrule{0em}{1pt}{1pt}
DynGEM & 2.0394  & 0.9623 & \textbf{0.4396} & 35.9251 \\
dynAE & 2.2982  & 2.1636  &  0.8245  & 33.6433 \\
dynRNN & 2.2563 & 1.3960  &  0.6634  &  37.9565  \\
dynAERNN & 1.8948 & 1.0784  & 0.7378 & 38.7106  \\
GCRN & 2.7111  & 4.1164 & 2.5884 & 122.3572 \\
EvolveGCN & 0.8963  & 4.6230 & 2.9725 & 104.3817 \\
CTGCN-C & 0.4258 & 4.1154 & 1.5013 & 13.4277 \\
CTGCN-S & 0.3693 & \textbf{0.3840} & 0.9029 & 22.7470 \\
\hline
\end{tabular}
\end{table}

We report the graph centrality prediction results for each dataset in Table \ref{table:5} and Table \ref{table:6}, respectively. We observe that the CGCN-S and CTGCN-S outperform baselines in the prediction accuracy of closeness, betweenness and k-core centrality. This indicates that the CTGCN-S is capable of preserving structural similarity between nodes in a global sense. We also note that the CGCN-C and CTGCN-C achieve outstanding centrality prediction performance. The possible reason is that the CGCN layers combine graph convolution operations in all k-cores, which enables structurally similar nodes in each k-core with similar representations. Hence, the results demonstrate that the CGCN layers are essential for capturing the global structure in dynamic graphs.

\subsection{Structural Role Classification}
We use a dynamic structural role classification task to further evaluate the structural similarity preserving ability of all the compared methods. In this task, the learned node representations are utilized to predict the structure role related labels of nodes in each dynamic graph. The classifier is logistic regression and the evaluation metric is accuracy. We evaluate the performance of all methods on two datasets: one is a European air-traffic network, and the other is an American air-traffic network \cite{ribeiro2017struc2vec}. As both datasets are static graphs, we randomly choose 10\% percent of the edges as the initial edges and add 10\% percent of the edges gradually to build 10 dynamic graph snapshots. Detailed statistics of both datasets are shown in Table \ref{table:7}.

\begin{table}[htbp]
\caption{Statistics of datasets utilized for structural role classification.}
\label{table:7}
\centering
\begin{tabular}{cccccc}
\hline
Dataset & \#Nodes & \#Edges & $D_{\rm max}$ & $k_{\rm max}$  & \#Snapshots \\
\hline
America-Air & 1190  & 13599 & 238 & 64 & 10 \\
Europe-Air  & 399  & 5995 & 202 & 33 & 10 \\
\hline
\end{tabular}
\end{table}

\begin{figure}[htbp]
\centering
\includegraphics[width=0.49\textwidth]{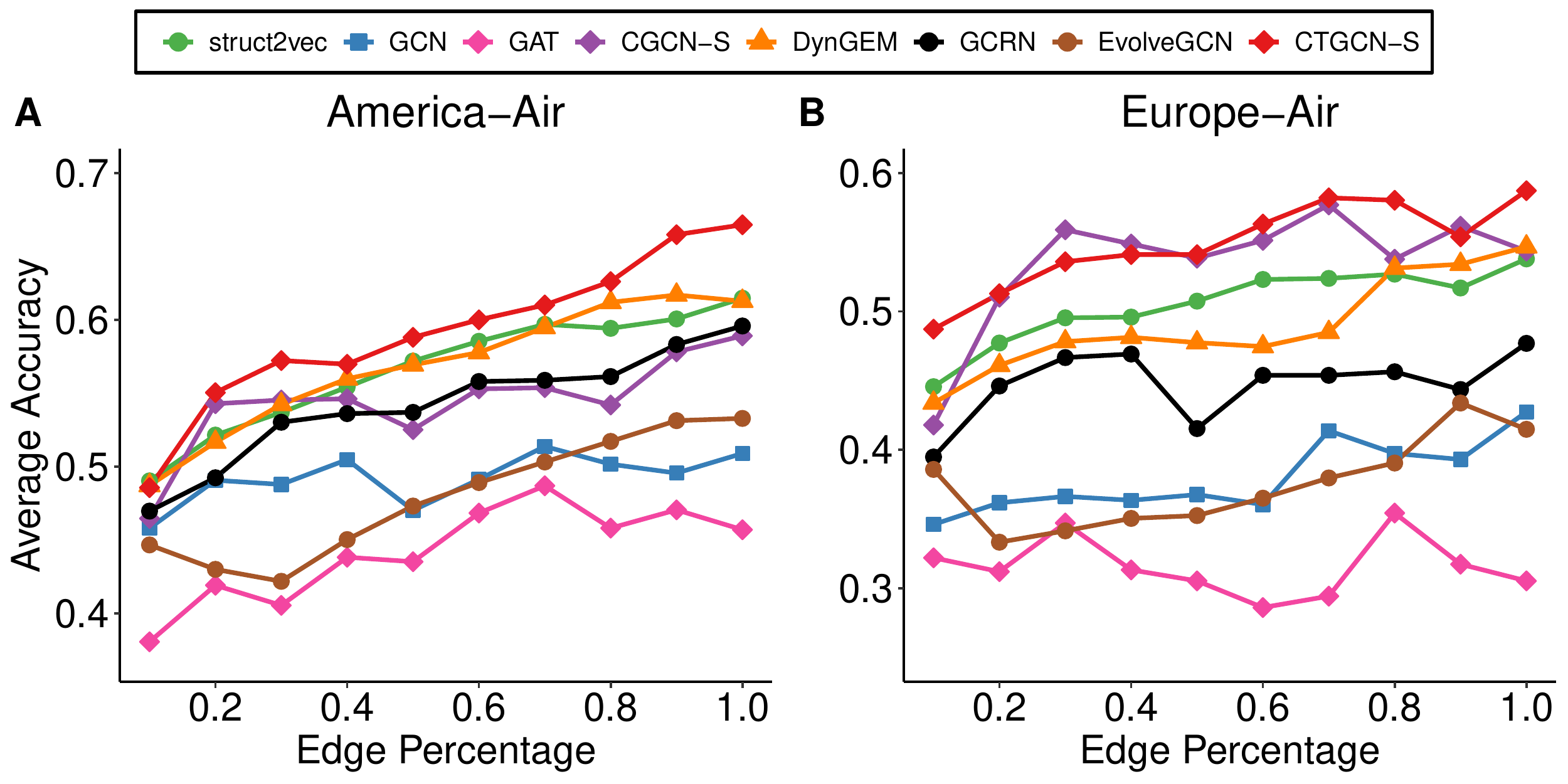}
\caption{Accuracy of structural role classification on dynamic graphs. }
\label{fig:5}
\end{figure}

\begin{figure*}[htbp]
\centering
\includegraphics[width=1\textwidth]{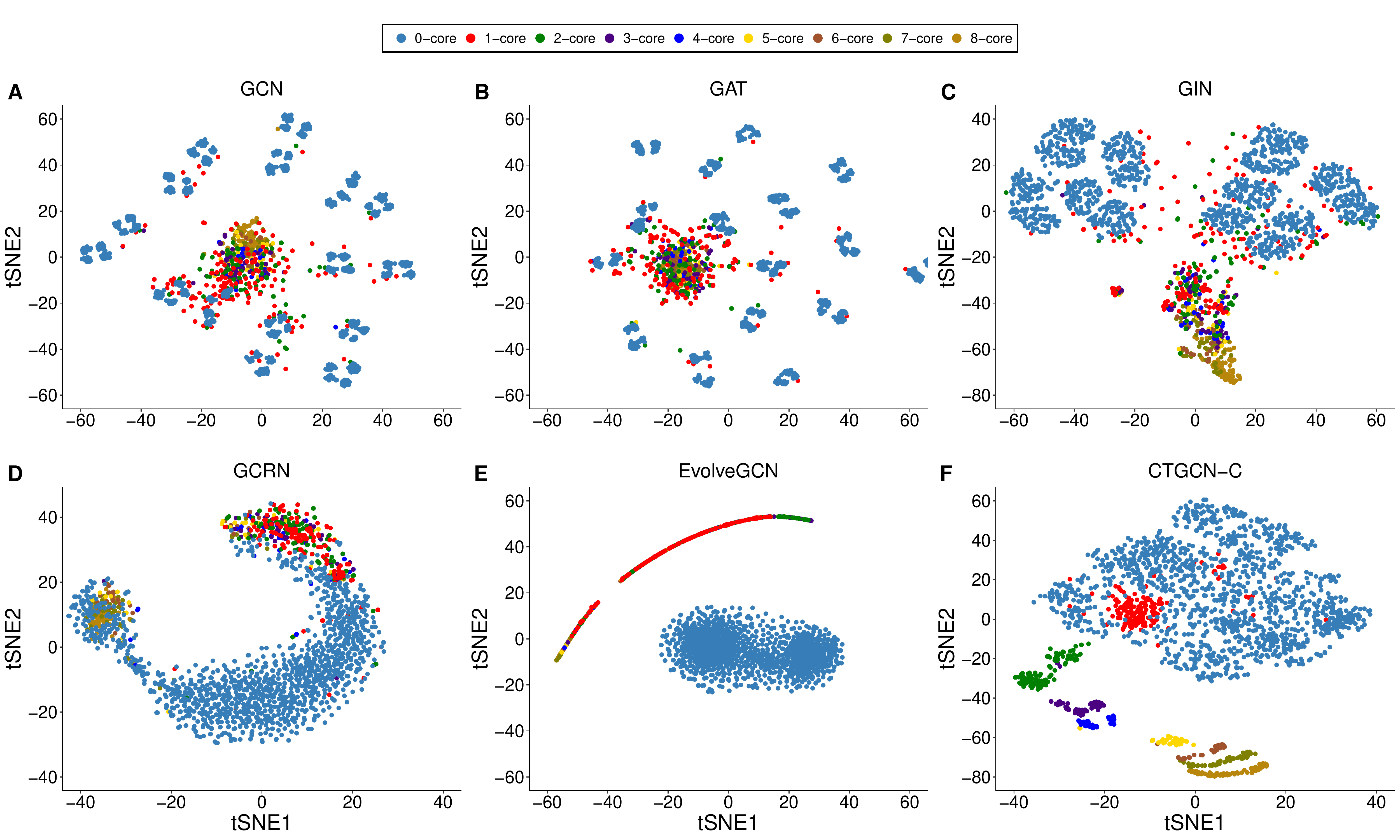}
\caption{2D t-SNE visualization of node embeddings on the UCI dataset. Different colors denote nodes in different cores. }
\label{fig:6}
\end{figure*}

The node classification results are shown in Figure \ref{fig:5}. We observe that our proposed CGCN-S and CTGCN-S methods outperform other baselines on both datasets, which indicates that the CTGCN-S can preserve global structural similarity between nodes in dynamic graphs. Furthermore, we note that struct2vec and DynGEM also achieve good structural role classification performance. This demonstrates that building a node-node structural similarity relationship, and encoding the neighboring information are effective ways to preserve graph structural information. As the CTGCN is an instance of encoding the neighboring information through CGCN layers, the results further demonstrate the effectiveness of our proposed feature aggregation architecture in CGCN layers.

\subsection{Visualization}
In the visualization task, we utilize t-distributed stochastic neighbor embedding 
(t-SNE) \cite{maaten2008visualizing}, a nonlinear dimensionality reduction and visualization approach, to project node embeddings of a dynamic graph into a 2D space. Our aim is to evaluate the global structure preserving ability of all GCN-based methods. The visualization results are shown in Figure \ref{fig:6}.

For static GCN-based methods, we can observe that in Figure \ref{fig:6}A-C, nodes in 0-core are agglomerated into several small clusters, and nodes in high-order cores are mixed together. This indicates that the GCN, GAT and GIN only preserve the local neighborhood information. As these methods do not take the global graph structure into consideration, they cannot capture the structural similarity among nodes in the same cores and the structural difference among nodes in different cores.

For dynamic GCN-based methods, although the GCRN and EvolveGCN cluster the 0-core nodes together, they still cannot distinguish the difference among nodes in high-order cores. In contrast to the GCRN and EvolveGCN, the CTGCN is capable of distinguishing all k-core subgraphs clearly, as shown in Figure \ref{fig:6}F. This demonstrates that the CTGCN not only preserves local connective proximity, but also captures global k-core subgraph information, which highly augments the expressive power of a GCN. Furthermore, this also indicates that the CTGCN is capable of capturing the latent k-core subgraph evolving trajectory in dynamic graphs.

\subsection{Ablation Study}
To further evaluate the effectiveness of the CTGCN, we first obtain two compared models. We replace the CGCN module with the vanilla GCN, that is, the GCRN; and simplify the CTGCN by omitting the RNN in each CGCN layer, namely, CTGCN-simple. Then, we perform detailed analyses on the link prediction performance of the GCRN, CTGCN-Simple and CTGCN.

\begin{table}[htbp]
\caption{Average AUC scores of all timestamps for ablation study.}
\label{table:8}
\centering
\begin{tabular}{ccccccc}
\hline
\specialrule{0em}{1pt}{1pt}
Method & UCI & AS & Math & Facebook & Enron \\
\hline
\specialrule{0em}{1pt}{1pt}
GCRN   &  0.8579   & 0.8648 &   0.8217  & 0.7262  &  0.8807  \\
CTGCN-Simple & 0.9042 &  0.9208   &  0.9523  &  0.8644  & 0.9637  \\
CTGCN-C &  \textbf{0.9434}  & 0.9578 & \textbf{0.9691} & \textbf{0.8836} & \textbf{0.9769}  \\
\hline
\end{tabular}
\end{table}

As illustrated in Table \ref{table:8}, the CTGCN-Simple consistently outperforms the GCRN on the above five datasets. This indicates that the k-core decomposition strategy is essential for the performance improvement of the CTGCN. We also observe that the CTGCN performs better than the CTGCN-Simple, which demonstrates that modeling the latent k-core subgraph evolution process by an RNN also helps improve the link prediction performance of the CTGCN.

\subsection{Parameter Sensitivity}
The proposed CTGCN methods involve a number of parameters that may affect its performance. We first examine how different choices of embedding dimension $d$ and the hyperparameter $Q$ in Equation \ref{eq:10} affect the link prediction performance of the CTGCN methods on the Facebook dataset. To save the computation time, we only test the link prediction performance on the last 5 snapshots.

\begin{figure}[htbp]
\centering
\includegraphics[width=0.49\textwidth]{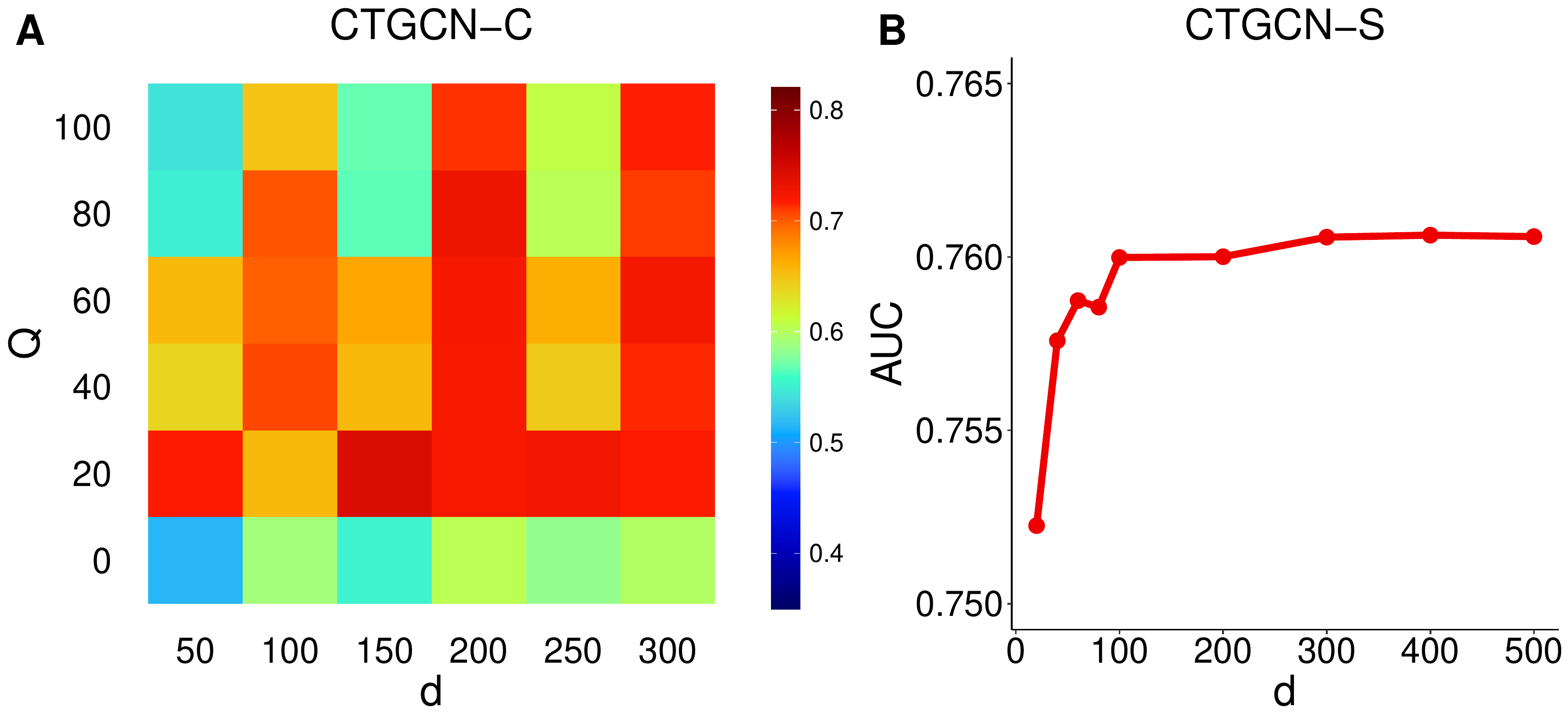}
\caption{Average AUC scores with respect to the hyperparameter $Q$ and the embedding dimension $d$ on the Facebook dataset. }
\label{fig:7}
\end{figure}

From the link prediction results shown in Figure \ref{fig:7}, we can see that the performance of the CTGCN-C increases when embedding dimension $d$ increases from 50 to 100, but tends to saturate when $d$ is greater than 100. Similarly, the performance of the CTGCN-S also becomes stable when embedding dimension $d$ reaches 100, but the CTGCN-S is more insensitive than the CTGCN-C with respect to the changes of $d$. We also observe that the CTGCN-C is sensitive with respect to the hyperparameter $Q$. The performance of the CTGCN-C increases and becomes stable when $Q$ reaches a certain value greater than 0, but drops when $Q$ is too large. This indicates that choosing an appropriate $Q$ value is helpful to the performance improvement of the CTGCN-C.

\begin{figure}[htbp]
\centering
\includegraphics[width=0.49\textwidth]{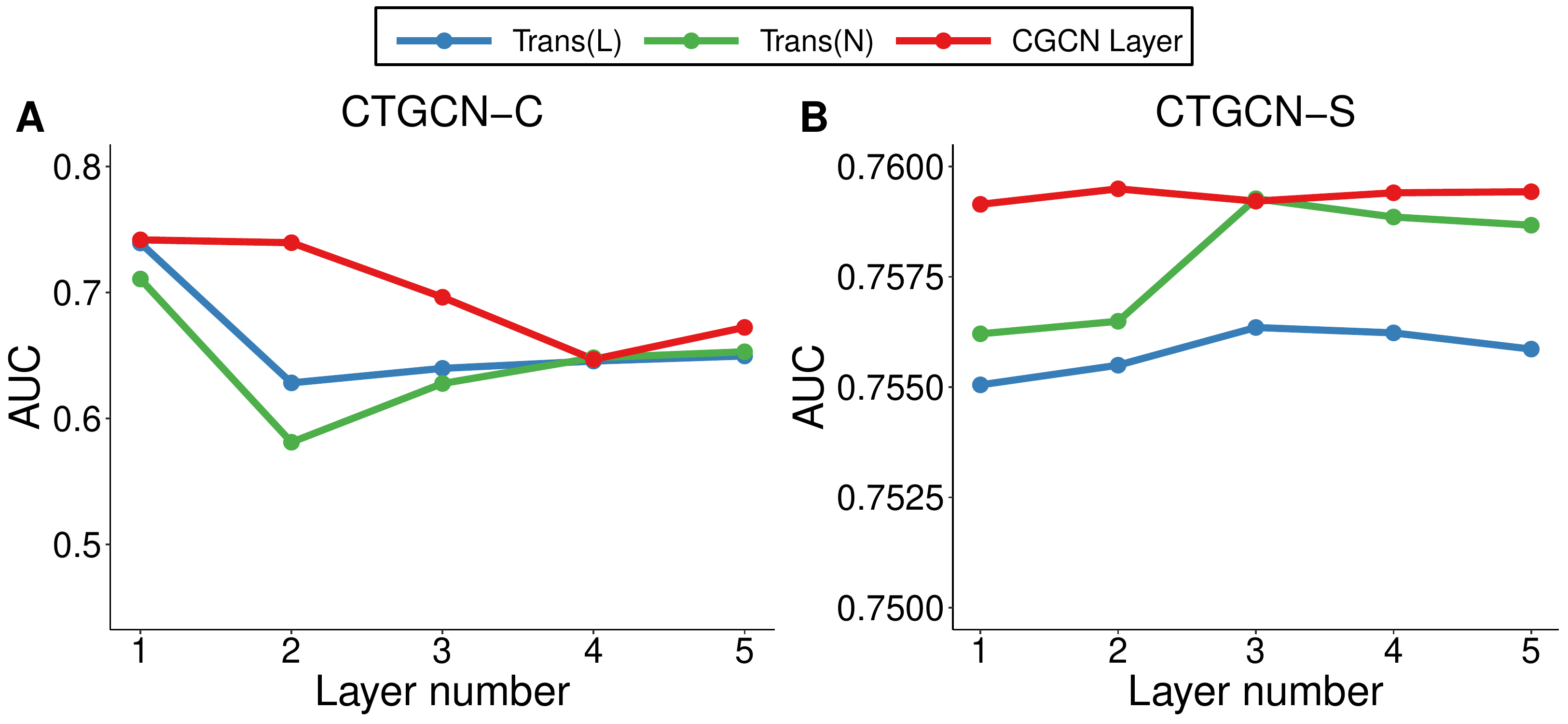}
\caption{Average AUC scores with respect to the number of transformation layers and CGCN layers on the Facebook dataset. Trans(L) is short for the linear MLP transformation layer, Trans(N) is short for the nonlinear MLP transformation layer.}
\label{fig:8}
\end{figure}

Then, we analyze the effect of the transformation layer number and CGCN layer number on the link prediction performance of the CTGCN-C and CTGCN-S. The experiments are conducted on the last 5 snapshots of the Facebook dataset. As illustrated in Figure \ref{fig:8}, we can see that the CTGCN-C is sensitive with respect to the transformation layer number and the CGCN layer number. The performance of the CTGCN-C drops when the transformation layer number or the CGCN layer number increases. This suggests that the CTGCN-C also encounters the oversmoothing problem when stacking too many graph convolution layers, similar to the GCN and GAT. In addition, the CTGCN-C is only compatible with 1-layer linear feature transformation. Utilizing either multilayer linear transformation or multilayer nonlinear transformation will result in the decay of its performance.

We also observe that the CTGCN-S is more robust with respect to the changes in layer numbers. When the CGCN layer number increases, the performance of the CTGCN-S is almost stable. When the transformation layer numbers reach 3, the performance of the CTGCN-S also becomes stable. For transformation layers, the CTGCN-S is more compatible with nonlinear transformation layers, as nonlinear transformation can yield more expressive node representations. It is worth noting that the CTGCN-S with either linear transformation layers or nonlinear transformation layers is comparable in the link prediction task, which suggests that the CTGCN-S can learn robust node representations in dynamic graphs.

\begin{figure}[htbp]
\centering
\includegraphics[width=0.49\textwidth]{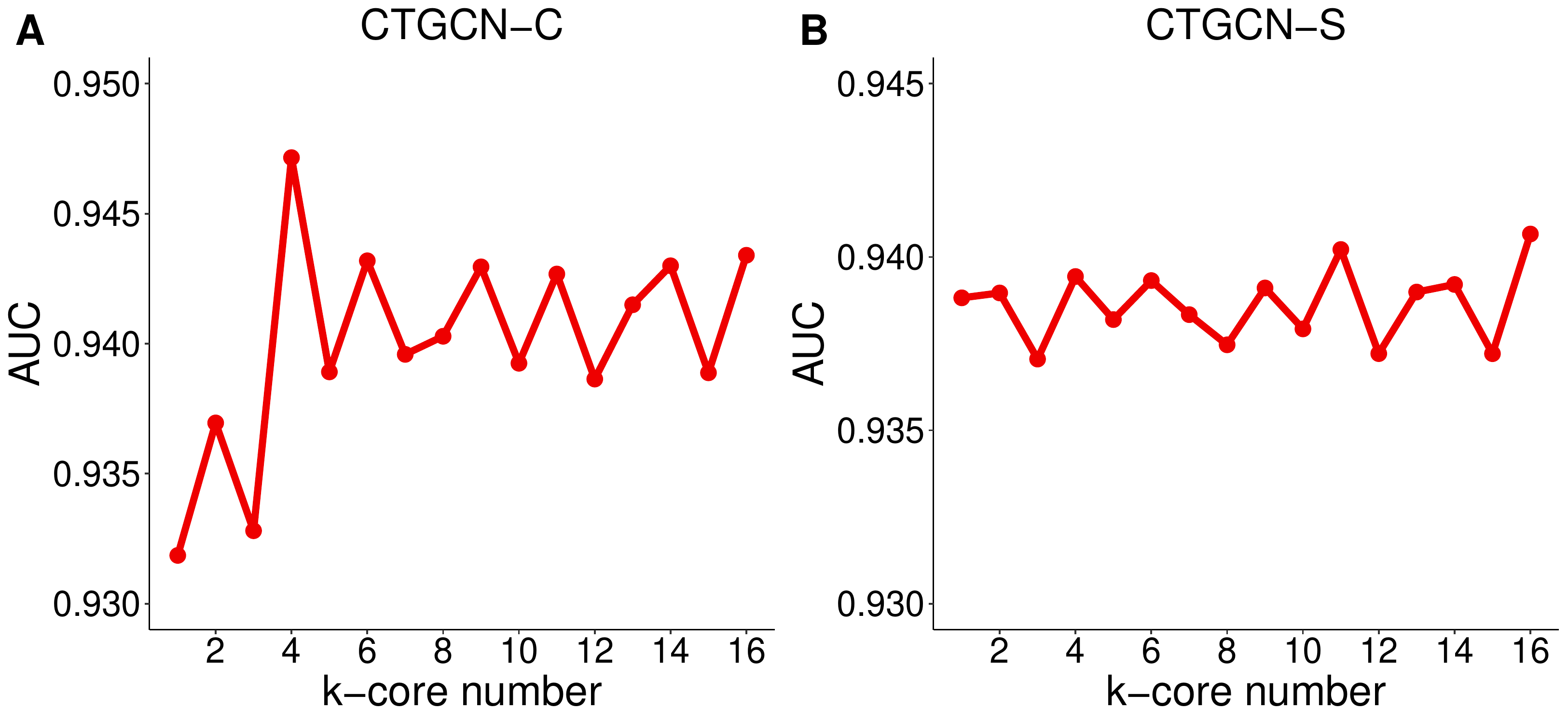}
\caption{Average AUC scores with respect to the k-core subgraph number in CGCN layers on the UCI dataset. }
\label{fig:9}
\end{figure}

Finally, we test the link prediction performance of the CTGCN methods with respect to the $k$-core subgraph number on the UCI dataset. The results are illustrated in Figure \ref{fig:9}. We observe that the link prediction performance of CTGCN-C increases when $k$ increases from 0 to 4, while the CTGCN-S does not show obvious performance improvement. The average AUC scores of both methods are nearly 0.940 when $k$ is large, which indicates that the CTGCN-C and CTGCN-S are insensitive to $k$. Hence, it is unnecessary to use all k-core subgraphs in practice. For large-scale dynamic graphs, we can use fewer k-core subgraphs in the CGCN layers to improve computational efficiency without reducing the link prediction performance.

\subsection{Efficiency Analysis}
In this section, we evaluate the efficiency of the proposed method from two aspects. First, we compare the running time of the CGCN-C and CGCN-S with those of static GCN-based methods. Second, we compare the running time of the CTGCN-C and CTGCN-S with those of dynamic graph embedding methods. Finally, we evaluate the scalability of the CTGCN-C and CTGCN-S with respect to the node number and the timestamp number.

\subsubsection{Running Time}
To make fair comparison, we enable GCN\footnote[1]{https://github.com/tkipf/pygcn}, GAT\footnote[2]{https://github.com/Diego999/pyGAT}, GIN\footnote[3]{https://github.com/weihua916/powerful-gnns} and the CGCN-C to have the same layer number and hidden dimensions. As the CGCN-S has a different architecture, we utilize the same CGCN-S settings as shown in Section \ref{exp_setup}. We set the k-core subgraph number as 5 for the CGCN-C and CGCN-S. All static methods are retrained on each timestamp of the Enron and Facebook datasets.

\begin{figure}[htbp]
\centering
\includegraphics[width=0.49\textwidth]{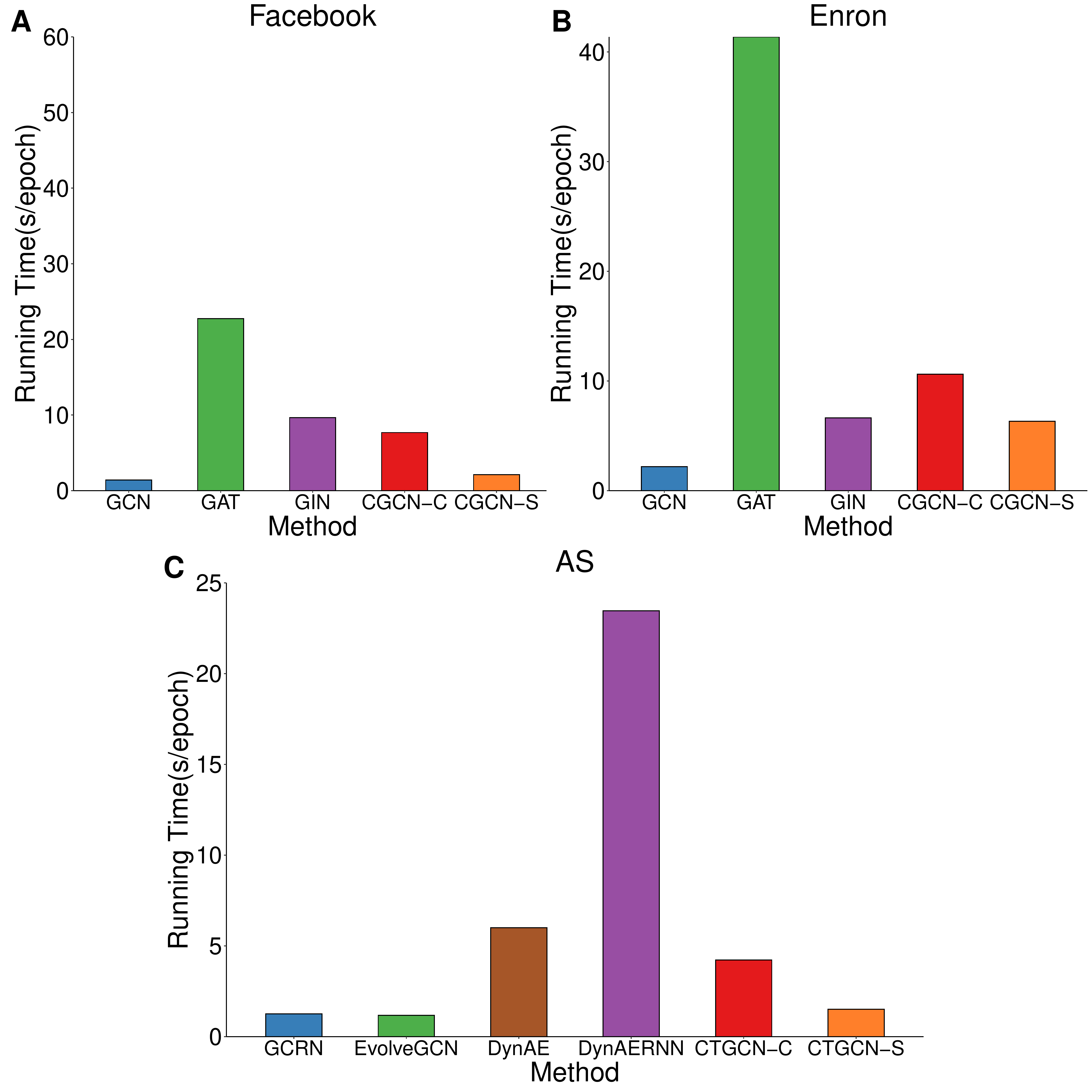}
\caption{Average training time of compared methods on all snapshots of the Facebook, Enron and AS datasets. }
\label{fig:10}
\end{figure}

Then, we count the average training time of all timestamps for each method. As illustrated in Figure \ref{fig:10}A-B, we can see that the GCN is highly efficient on both large-scale datasets. The CGCN-C and CGCN-S are as efficient as the GIN and faster than the GAT. The possible reason is that the attention operation is time-consuming on large graphs, while the proposed CGCN-C and CGCN-S share the feature transformation matrix across all k-core adjacency matrices, which highly reduces the time complexity. Therefore, the results demonstrate that the CGCN-C and CGCN-S are efficient on large-scale dynamic graphs.

After the running time comparison of static GCN methods, we then compare the running time of all compared dynamic graph embedding methods. We use 5 dynamic graphs of the AS dataset as input, and set the k-core subgraph number as 5 for the CTGCN-C and CTGCN-S. The results are shown in Figure \ref{fig:10}C. We can see that the CTGCN-S is as efficient as the GCRN and EvolveGCN, and the CTGCN-C is faster than dyngraph2vec variants. This demonstrates that our proposed CTGCN is computationally efficient and can be applied to large scale dynamic graphs. 

\subsubsection{Scalability}
We count the actual running time per epoch for the CTGCN-C and CTGCN-S on the Facebook dataset to test their scalability. We sample subgraphs with different node numbers varying from 100 to 60730 on a graph to test the scalability with respect to the node number, and train models on different timestamp ranges to test the scalability with respect to the timestamp number. The experiments are run on a machine with 48 Intel Xeon Gold 2.30GHz CPU processors and 256G RAM. The CTGCN-C and CTGCN-S settings are the same as those in Section \ref{exp_setup}. 

\begin{figure}[htbp]
\centering
\includegraphics[width=0.49\textwidth]{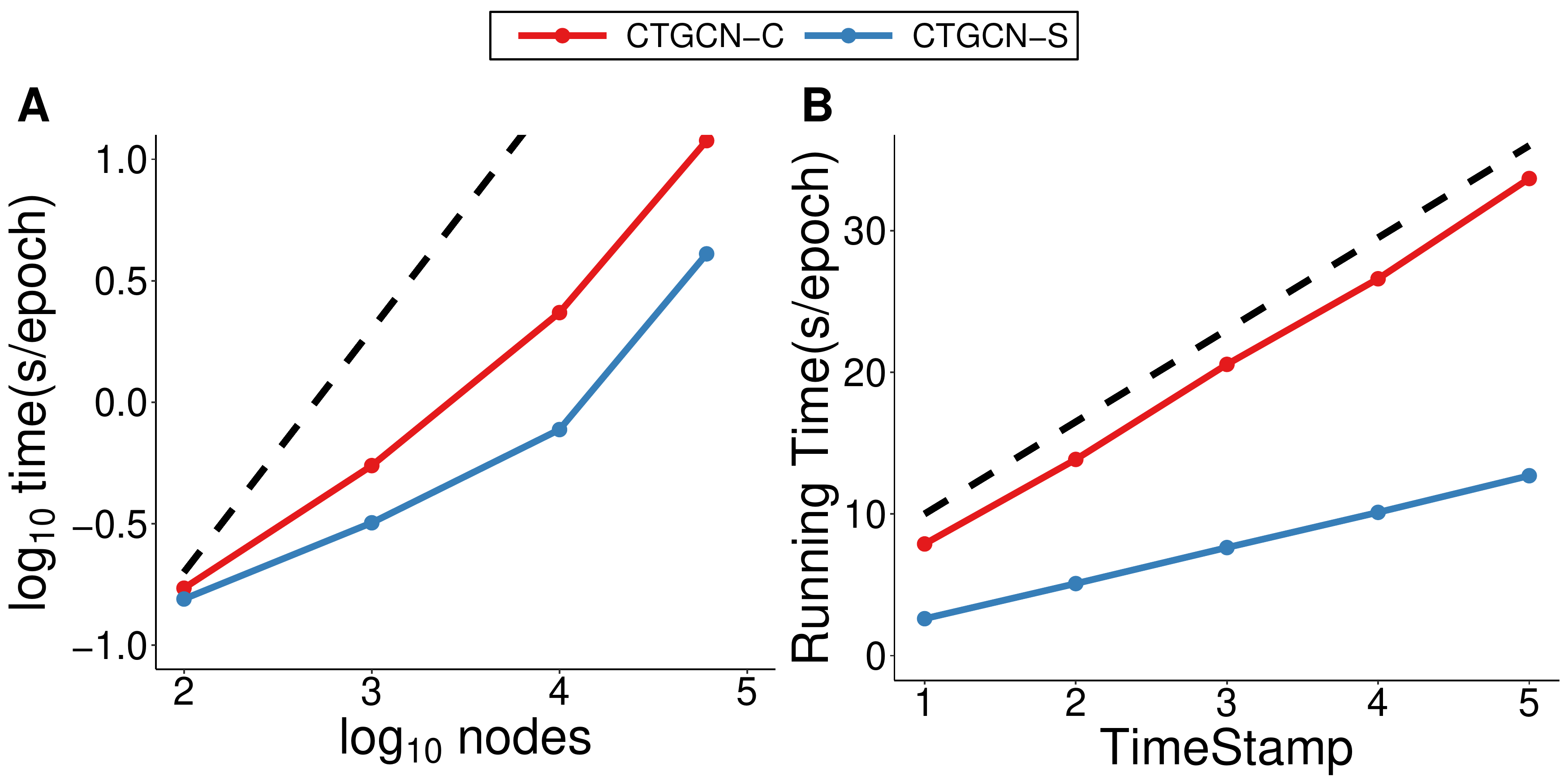}
\caption{Training time of CTGCNs with respect to the node number and the timestamp number on the Facebook dataset. }
\label{fig:11}
\end{figure}

As shown in Figure \ref{fig:11}, we can observe that the training time of the CTGCN-C scales linearly with the number of nodes and the slope of the curve is close to 1. The training time of the CTGCN-S also scales linearly with the node number, and the CTGCN-S runs faster than the CTGCN-C. Furthermore, we can see that both the CTGCN-C and CTGCN-S scale linearly with the timestamp number. The results are consistent with the complexity analysis in Section \ref{comp_ana} and demonstrate that our methods are scalable on large-scale dynamic graphs.

\section{Conclusion}
In this paper, we propose a novel k-core based temporal GCN(CTGCN) for dynamic graph embedding. This method can preserve both connective proximity and structural similarity in a unified generalized GCN framework. In the CTGCN, we present a novel core-based graph convolutional layer that can capture hierarchical structure information in nested k-core subgraphs. Experimental results on several real-world graphs demonstrate the effectiveness and efficiency of the proposed method. In contrast to previous GCNs, the CTGCN is essentially an instance of our proposed generalized GCN framework which regards graph convolution as feature transformation and feature aggregation phases. Our future work will develop novel GCN models under the proposed generalized GCN framework.

\ifCLASSOPTIONcompsoc
  \section*{Acknowledgments}
\else
  \section*{Acknowledgment}
\fi
This work is supported by HuaRong RongTong (Beijing) Technology Co., Ltd. We acknowledge HuaRong RongTong(Beijing) for providing us high-performance machines for computation. We thank Yusong Ye, Yunchang Zhu, Xiang Li for discussion. We also acknowledge anonymous reviewers for proposing detailed modification advice to help us improve the quality of this manuscript.

\ifCLASSOPTIONcaptionsoff
  \newpage
\fi



\bibliographystyle{IEEEtran}
\end{document}